\documentclass[11pt,a4paper]{article}
\usepackage[hyperref]{acl2017}
\usepackage{times}
\usepackage{latexsym}
\usepackage{amsmath}
\usepackage{graphicx}
\usepackage{url}

\aclfinalcopy 

\title{Weakly Supervised Cross-Lingual Named Entity Recognition\\via Effective Annotation and Representation Projection}

\author{Jian Ni \and Georgiana Dinu \and Radu Florian\\
    IBM T. J. Watson Research Center\\
    1101 Kitchawan Road, Yorktown Heights, NY 10598, USA\\
  {\tt \{nij, gdinu, raduf\}@us.ibm.com}}

\date{}

\begin{document}
\maketitle

\begin{abstract}
The state-of-the-art named entity recognition (NER) systems are supervised machine learning models that require large amounts of manually annotated data to achieve high accuracy. However, annotating NER data by human is expensive and time-consuming, and can be quite difficult for a new language. In this paper, we present two weakly supervised approaches for cross-lingual NER with no human annotation in a target language. The first approach is to create automatically labeled NER data for a target language via annotation projection on comparable corpora, where we develop a heuristic scheme that effectively selects good-quality projection-labeled data from noisy data. The second approach is to project distributed representations of words (word embeddings) from a target language to a source language, so that the source-language NER system can be applied to the target language without re-training. We also design two co-decoding schemes that effectively combine the outputs of the two projection-based approaches. We evaluate the performance of the proposed approaches on both in-house and open NER data for several target languages. The results show that the combined systems outperform three other weakly supervised approaches on the CoNLL data.
\end{abstract}

\section{Introduction}

Named entity recognition (NER) is a fundamental information extraction task that automatically detects named entities in text and classifies them into pre-defined entity types such as PERSON, ORGANIZATION, GPE (GeoPolitical Entities), EVENT, LOCATION, TIME, DATE, etc. NER provides essential inputs for many information extraction applications, including relation extraction, entity linking, question answering and text mining. Building fast and accurate NER systems is a crucial step towards enabling large-scale automated information extraction and knowledge discovery on the huge volumes of electronic documents existing today.

The state-of-the-art NER systems are supervised machine learning models \cite{nadeau07}, including maximum entropy Markov models (MEMMs) \cite{mccallum00}, conditional random fields (CRFs) \cite{lafferty01} and neural networks \cite{collobert11,lample16}. To achieve high accuracy, a NER system needs to be trained with a large amount of manually annotated data, and is often supplied with language-specific resources (e.g., gazetteers, word clusters, etc.). Annotating NER data by human is rather expensive and time-consuming, and can be quite difficult for a new language. This creates a big challenge in building NER systems of multiple languages for supporting multilingual information extraction applications.

The difficulty of acquiring supervised annotation raises the following question: given a well-trained NER system in a source language (e.g., English), how can one go about extending it to a new language with decent performance and no human annotation in the target language? There are mainly two types of approaches for building weakly supervised cross-lingual NER systems. 

The first type of approaches create weakly labeled NER training data in a target language. One way to create weakly labeled data is through annotation projection on aligned parallel corpora or translations between a source language and a target language, e.g., \cite{yarowsky01,zitouni08,ehrmann11}. Another way is to utilize the text and structure of Wikipedia to generate weakly labeled multilingual training annotations, e.g., \cite{richman08,nothman13,alrfou15}.

The second type of approaches are based on direct model transfer, e.g.,  \cite{tackstrom12,tsai16}. The basic idea is to train a single
NER system in the source language with language independent features, so the system can be applied to other languages using those universal features.

In this paper, we make the following contributions to weakly supervised cross-lingual NER with no human annotation in the target languages. First, for the \textit{annotation projection} approach, we develop a heuristic, language-independent data selection scheme that seeks to select good-quality projection-labeled NER data from comparable corpora. Experimental results show that the data selection scheme can significantly improve the accuracy of the target-language NER system when the alignment quality is low and the projection-labeled data are noisy.

Second, we propose a new approach for direct NER model transfer based on \textit{representation projection}. It projects word representations in vector space (word embeddings) from a target language to a source language, to create a universal representation of the words in different languages. Under this approach, the NER system trained for the source language can be directly applied to the target language without the need for re-training.

Finally, we design two \textit{co-decoding} schemes that combine the outputs (views) of the two projection-based systems to produce an output that is more accurate than the outputs of individual systems.  We evaluate the performance of the proposed approaches on both in-house and open NER data sets for a number of target languages. The results show that the combined systems outperform the state-of-the-art cross-lingual NER approaches proposed in \newcite{tackstrom12}, \newcite{nothman13} and \newcite{tsai16} on the CoNLL NER test data \cite{sang02,sang03}.

We organize the paper as follows. In Section 2 we introduce three NER models that are used in the paper. In Section 3 we present an annotation projection approach with effective data selection. In Section 4 we propose a representation projection approach for direct NER model transfer. In Section 5 we describe two co-decoding schemes that effectively combine the outputs of two projection-based approaches. In Section 6 we evaluate the performance of the proposed approaches. We describe related work in Section 7 and conclude the paper in Section 8.

\section{NER Models}

The NER task can be formulated as a sequence labeling problem: given a sequence of words $x_1,...,x_n$, we want to infer the NER tag $l_i$ for each word $x_i$, $1\leq i \leq n$. In this section we introduce three NER models that are used in the paper.

\subsection{CRFs and MEMMs}

\emph{Conditional random fields} (CRFs) are a class of discriminative probabilistic graphical models that provide powerful tools for labeling sequential data \cite{lafferty01}. CRFs learn a conditional probability model $p_\lambda(\mathbf{l}|\mathbf{x})$ from a set of labeled training data, where $\mathbf{x}=(\mathbf{x}_1,...,\mathbf{x}_n)$ is a random sequence of input words, $\mathbf{l}=(\mathbf{l}_1,...,\mathbf{l}_n)$ is the sequence of label variables (NER tags) for $\mathbf{x}$, and $\mathbf{l}$ has certain Markov properties conditioned on $\mathbf{x}$. Specifically, a general-order CRF with order $o$ assumes that label variable $\mathbf{l}_i$ is dependent on a fixed number $o$ of previous label variables $\mathbf{l}_{i-1},...,\mathbf{l}_{i-o}$, with the following conditional distribution:
\begin{equation} \label{eq:general-order-crf}
p_{\lambda}(\mathbf{l | x})  = \frac{e^{\sum_{i=1}^{n} \sum_{k=1}^{K}\lambda_k f_k(\mathbf{l}_{i},\mathbf{l}_{i-1},...,\mathbf{l}_{i-o},\mathbf{x})}}{Z_{\lambda}(\mathbf{x})}
\end{equation}
where $f_k$'s are feature functions, $\lambda_k$'s are weights of the feature functions (parameters to learn), and $Z_{\lambda}(\mathbf{x})$ is a normalization constant. When $o=1$, we have a first-order CRF which is also known as a linear-chain CRF.

Given a set of labeled training data $\mathcal{D}=(\mathbf{x}^{(j)},\mathbf{l}^{(j)})_{j=1,...,N}$, we seek to find an optimal set of parameters $\lambda^*$ that maximize the conditional log-likelihood of the data:
\begin{equation}
\lambda^* = \arg\max_{\lambda} \sum_{j=1}^{N} \log p_{\lambda}(\mathbf{l}^{(j)}|\mathbf{x}^{(j)})
\end{equation}
Once we obtain $\lambda^*$, we can use the trained model $p_{\lambda^*}(\mathbf{l|x})$ to decode the most likely label sequence
$\mathbf{l}^*$ for any new input sequence of words $\mathbf{x}$ (via the Viterbi algorithm for example):
\begin{equation}
\mathbf{l}^* = \arg\max_{\mathbf{l}}p_{\lambda^*}(\mathbf{l|x})
\end{equation}

A related conditional probability model, called \emph{maximum entropy Markov model} (MEMM) \cite{mccallum00}, assumes that $\mathbf{l}$ is a Markov chain conditioned on $\mathbf{x}$:
\begin{eqnarray} \label{eq:general-order-memm}
p_{\lambda}(\mathbf{l|x}) & = & \prod_{i=1}^{n} p_{\lambda}(\mathbf{l}_{i} | \mathbf{l}_{i-1}, ..., \mathbf{l}_{i-o}, \mathbf{x}) \nonumber \\
& = & \prod_{i=1}^{n} \frac{e^{\sum_{k=1}^{K} \lambda_k f_k(\mathbf{l}_i,\mathbf{l}_{i-1},...,\mathbf{l}_{i-o},
\mathbf{x})}}{Z_{\lambda}(\mathbf{l}_{i-1},...,\mathbf{l}_{i-o}, \mathbf{x})} 
\end{eqnarray}

The main difference between CRFs and MEMMs is that CRFs normalize the conditional distribution over the whole sequence as in (\ref{eq:general-order-crf}), while MEMMs normalize the conditional distribution per token as in (\ref{eq:general-order-memm}). As a result, CRFs can better handle the label bias problem \cite{lafferty01}. This benefit, however, comes at a price. The training time of order-$o$ CRFs grows exponentially ($O(M^{o+1})$) with the number of output labels $M$, which is typically slow even for moderate-size training data if $M$ is large. In contrast, the training time of order-$o$ MEMMs is linear ($O(M)$) with respect to $M$ independent of $o$, so it can handle larger training data with higher order of dependency. We have implemented both a linear-chain CRF model and a general-order MEMM model.

\subsection{Neural Networks}

With the increasing popularity of distributed (vector) representations of words, neural network models have recently been applied to tackle many NLP tasks including NER \cite{collobert11,lample16}.

We have implemented a feedforward neural network model which maximizes the log-likelihood of the training data similar to that of \cite{collobert11}. We adopt a locally normalized model (the conditional distribution is normalized per token as in MEMMs) and introduce context dependency by conditioning on the previously assigned tags. We use a target word and its surrounding context as features. We do not use other common features such as gazetteers or character-level representations as such features might not be readily available or might not transfer to other languages. 

\begin{figure}
\includegraphics[scale=0.105]{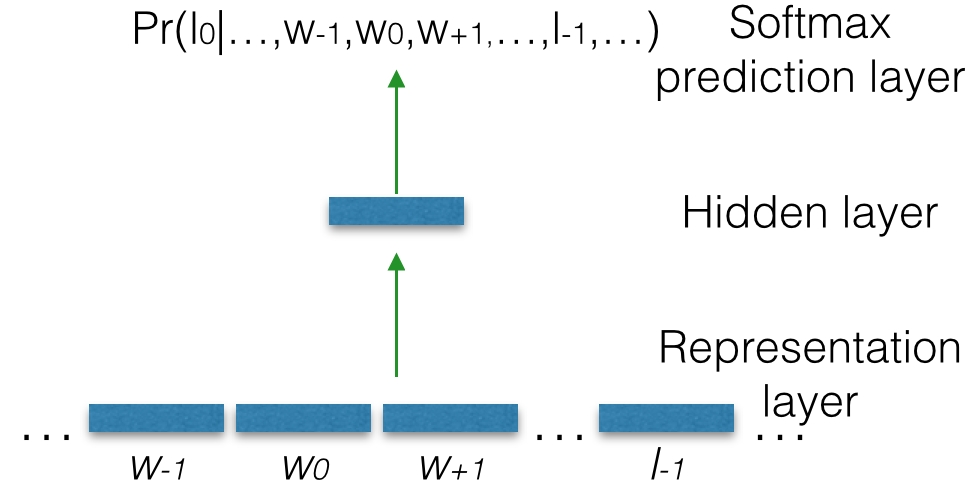}
\includegraphics[scale=0.105]{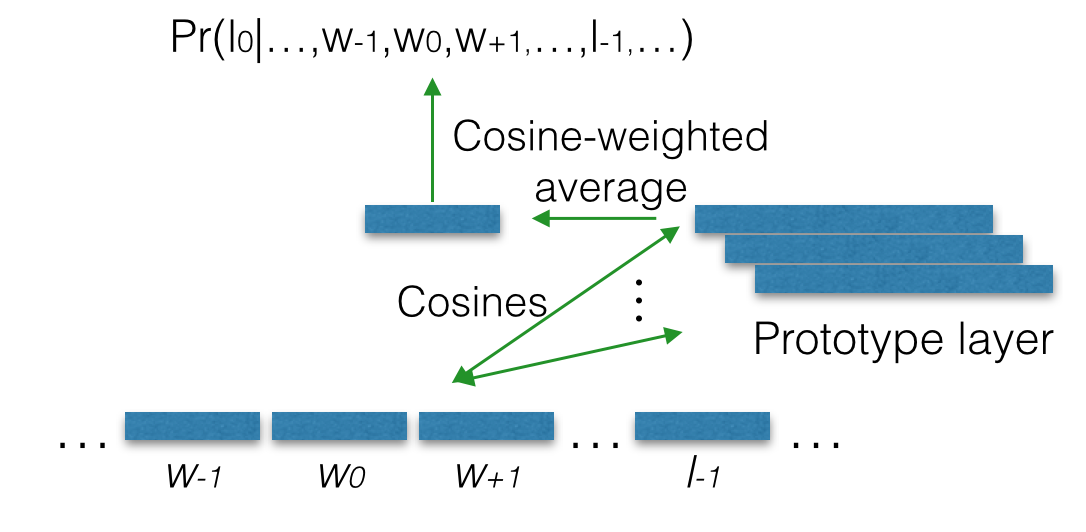}
\caption{Architecture of the two neural network models: left-NN1, right-NN2.}
\label{figure:nn-arch}
\end{figure}

We have deployed two neural network architectures. The first one (called NN1) uses the word embedding of a word as the input. The second one (called NN2) adds a smoothing prototype layer that computes the cosine similarity between a word embedding and a fixed set of prototype vectors (learned during training) and returns a weighted average of these prototype vectors as the input. In our experiments we find that with the smoothing layer, NN2 tends to have a more balanced precision and recall than NN1. Both networks have one hidden layer, with sigmoid and softmax activation functions on the hidden and output layers respectively. The two neural network models are depicted in Figure \ref{figure:nn-arch}.

\section{Annotation Projection Approach}

The existing annotation projection approaches require parallel corpora or translations between a source language and a target language with alignment information. In this paper, we develop a heuristic, language-independent data selection scheme that seeks to select good-quality projection-labeled data from noisy comparable corpora. We use English as the source language. 

Suppose we have comparable\footnote{Ideally, the sentences would be translations of each other, but we only require possibly parallel sentences.} sentence pairs $(\mathbf{X},\mathbf{Y})$ between English and a target language, where $\textbf{X}$ includes $N$ English sentences $\mathbf{x}^{(1)},...,\mathbf{x}^{(N)}$, $\textbf{Y}$ includes $N$ target-language sentences $\mathbf{y}^{(1)},...,\mathbf{y}^{(N)}$, and $\mathbf{y}^{(j)}$ is aligned to $\mathbf{x}^{(j)}$ via an alignment model, $1\leq j \leq N$. We use a sentence pair ($\mathbf{x}, \mathbf{y})$ as an example to illustrate how the annotation projection procedure works, where $\mathbf{x}=(x_1,x_2,...,x_s)$ is an English sentence, and $\mathbf{y}=(y_1,y_2,...,y_t)$ is a target-language sentence that is aligned to $\mathbf{x}$.\\

\textit{Annotation Projection Procedure}
\begin{itemize}
\item[1.] Apply the English NER system on the English sentence $\mathbf{x}$ to generate the NER tags $\mathbf{l}=(l_1,l_2,...,l_s)$ for $\mathbf{x}$.

\item[2.] Project the NER tags to the target-language sentence $\mathbf{y}$ using the alignment information. Specifically, if a sequence
of English words $(x_i,...,x_{i+p})$ is aligned to a sequence of target-language words $(y_j,...,y_{j+q})$, and $(x_i,...,x_{i+p})$ is
recognized (by the English NER system) as an entity with NER tag $l$, then $(y_j,...,y_{j+q})$ is labeled with $l$\footnote{If the IOB (Inside, Outside, Beginning) tagging format  is used, then $(y_j,y_{j+1},...,y_{j+q})$ is labeled with (B-$l$, I-$l$,...,I-$l$).}. \\
Let $\mathbf{l}'=(l'_1,l'_2,...,l'_t)$ be the projected NER tags for the target-language sentence $\mathbf{y}$.
\end{itemize}

We can apply the annotation projection procedure on all the sentence pairs $(\mathbf{X},\mathbf{Y})$, to generate projected NER tags $\mathbf{L}'$ for the target-language sentences $\mathbf{Y}$. ($\mathbf{Y}$, $\mathbf{L}'$) are automatically labeled NER data with no human annotation in the target language. One can use those projection-labeled data to train an NER system in the target language. The quality of such weakly labeled NER data, and consequently the accuracy of the target-language NER system, depend on both 1) the accuracy of the English NER system, and 2) the alignment accuracy of the sentence pairs.

Since we don't require actual translations, but only comparable data, the downside is that if some of the data are not actually parallel and if we use all for weakly supervised learning, the accuracy of the target-language NER system might be adversely affected. We are therefore motivated to design effective data selection schemes that can select good-quality projection-labeled data from noisy data, to improve the accuracy of the annotation projection approach for cross-lingual NER.

\subsection{Data Selection Scheme}

We first design a metric to measure the annotation quality of a projection-labeled sentence in the target language. We construct a frequency table $\mathrm{T}$ which includes all the entities in the projection-labeled target-language sentences. For each entity $e$, $\mathrm{T}$ also includes the projected NER tags for $e$ and the relative frequency (empirical probability) $\hat{P}(l|e)$ that entity $e$ is labeled with tag $l$. Table \ref{table:frequency-table} shows a snapshot of the frequency table where the target language is Portuguese.

We use $\hat{P}(l|e)$ to measure the reliability of labeling entity $e$ with tag $l$ in the target language. The intuition is that if an entity $e$ is labeled by a tag $l$ with higher frequency than other tags in the projection-labeled data, it is more likely that the annotation is correct. For example, if the joint accuracy of the source NER system and alignment system is greater than 0.5, then the correct tag of a random entity will have a higher relative frequency than other tags in a large enough sample.

Based on the frequency scores, we calculate the quality score of a projection-labeled target-language sentence $\mathbf{y}$ by
averaging the frequency scores of the projected entities in the sentence:
\begin{equation}
q(\mathbf{y})  =  \frac{\Sigma_{e\in \mathbf{y}} \hat{P}(l'(e)|e)}{n(\mathbf{y})}
\end{equation}
where $l'(e)$ is the projected NER tag for $e$, and $n(\mathbf{y})$ is the total number of entities in sentence $\mathbf{y}$.

We use $q(\mathbf{y})$ to measure the annotation quality of sentence $\mathbf{y}$, and $n(\mathbf{y})$ to measure the amount of annotation information contained in sentence $\mathbf{y}$. We design a \emph{heuristic data selection scheme} which selects projection-labeled sentences in the target language that satisfy the following condition:
\begin{equation}
q(\mathbf{y})  \geq q; \mbox{   } n(\mathbf{y}) \geq n
\end{equation}
where $q$ is a quality score threshold and $n$ is an entity number threshold. We can tune the two parameters to make tradeoffs among the annotation quality of the selected sentences, the annotation information contained in the selected sentences, and the total number of sentence selected.

One way to select the threshold parameters $q$ and $n$ is via a development set - either a small set of human-annotated data or a sample of the projection-labeled data. We select the threshold parameters via \textit{coordinate search} using the development set: we first fix $n=3$ and search the best $\hat{q}$ in $[0,0.9]$ with a step size of 0.1; we then fix $q=\hat{q}$ and select the best $\hat{n}$ in $[1,5]$ with a step size of 1.

\begin{table}
\small \centering
\begin{center}
\begin{tabular}{|c|c|c|}

\hline \textbf{Entity Name} & \textbf{NER Tag} & \textbf{Frequency}  \\
\hline Estados Unidos & GPE & 0.853 \\
\hline Estados Unidos & ORGANIZATION & 0.143 \\
\hline Estados Unidos & PEOPLE & 0.001 \\
\hline Estados Unidos & PRODUCT & 0.001 \\
\hline Estados Unidos & TITLEWORK & 0.001 \\
\hline Estados Unidos & EVENT & 0.001 \\
\hline
\end{tabular}
\end{center}
\caption{A snapshot of the frequency table where the target language is Portuguese. \textit{Estados Unidos} means \textit{United States}. The correct NER tag for Estados Unidos is GPE which has the highest relative frequency in the weakly labeled data.}
\label{table:frequency-table}
\end{table}

\subsection{Accuracy Improvements}

We evaluate the effectiveness of the data selection scheme via experiments on 4 target languages: Japanese, Korean, German and Portuguese. We use comparable corpora between English and each target language (ranging from 2M to 6M tokens) with alignment information.  For each target language, we also have a set of manually annotated NER data (ranging from 30K to 45K tokens) which are served as the test data for evaluating the target-language NER system. 

The source (English) NER system is a linear-chain CRF model which achieves an accuracy of 88.9 $F_1$ score on an independent NER test set. The alignment systems between English and the target languages are maximum entropy models \cite{ittycheriah05}, with an accuracy of 69.4/62.0/76.1/88.0 $F_1$ score on independent Japanese/Korean/German/Portuguese alignment test sets.

For each target language, we randomly select 5\% of the projection-labeled data as the development set and the remaining 95\% as the training set. We compare an NER system trained with all the projection-labeled training data with no data selection (i.e., $(q,n)=(0,0)$) and an NER system trained with projection-labeled data selected by the data selection scheme where the development set is used to select the threshold parameters $q$ and $n$ via coordinate search.  Both NER systems are 2nd-order MEMM models\footnote{In our experiments, CRFs cannot handle training data with a few million words, since our NER system has over 50 entity types, and the training time of CRFs grows at least quadratically in the number of entity types.} which use the same template of features.

\begin{table}
\small\centering
\begin{center}
\begin{tabular}{|c|c|c|c|}

\hline \textbf{Language} & $(q,n)$ & \textbf{Training Size} & \textbf{$F_1$ Score} \\
\hline Japanese & (0, 0) & 4.9M & 41.2\\
\cline{2-4}  & (0.7, 4) & 1.3M  & 53.4\\
\hline Korean & (0, 0) & 4.5M & 25.0\\
\cline{2-4}  & (0.4, 2) & 1.5M & 38.7 \\
\hline German & (0, 0) & 5.2M & 67.2\\
\cline{2-4}  & (0.4, 4) & 2.6M & 67.5\\
\hline Portuguese & (0, 0) & 2.1M & 61.5\\
\cline{2-4}  &  (0.1, 4) & 1.5M & 62.7\\
\hline
\end{tabular}
\end{center}
\caption{Performance comparison of weakly supervised NER systems trained without data selection ($(q,n)=(0,0)$) and with data selection ($(\hat{q},\hat{n})$ determined by coordinate search).} \label{table:data-selection-improvement}
\end{table}

The results are shown in Table \ref{table:data-selection-improvement}. For different target languages, we use the same source (English) NER system for annotation projection, so the differences in the accuracy improvements are mainly due to the alignment quality of the comparable corpora between English and different target languages. When the alignment quality is low (e.g., as for Japanese and Korean) and hence the projection-labeled NER data are quite noisy, the proposed data selection scheme is very effective in selecting good-quality projection-labeled data and the improvement is big: +12.2 $F_1$ score for Japanese and +13.7 $F_1$ score for Korean. Using a stratified shuffling test \cite{noreen88}, for a significance level of 0.05, data-selection is statistically significantly better than no-selection for Japanese, Korean and Portuguese.

\section{Representation Projection Approach}

In this paper, we propose a new approach for direct NER model transfer based on representation projection. Under this approach, we train a single English NER system that uses only word embeddings as input representations. We create mapping functions which can map words in any language into English and we simply use the English NER system to decode. In particular, by mapping all languages into English, we are using one universal NER system and we do not need to re-train the system when a new language is added.

\subsection{Monolingual Word Embeddings}

We first build vector representations of words (word embeddings) for a language using monolingual data. We use a variant of the Continuous Bag-of-Words (CBOW) word2vec model \cite{mikolov13}, which concatenates the context words surrounding a target word instead of adding them (similarly to \cite{ling15}). Additionally, we employ weights $w=\frac{1}{dist(x,x_c)}$ that decay with the distance of a context word $x_c$ to a target word $x$. Tests on word similarity benchmarks show this variant leads to small improvements over the standard CBOW model. 

We train 300-dimensional word embeddings for English. Following  \cite{mikolov13b}, we use larger dimensional embeddings for the target languages, namely 800. We train word2vec for 1 epoch for English/Spanish and 5 epochs for the rest of the languages for which we have less data.

\subsection{Cross-Lingual Representation Projection}

We learn cross-lingual word embedding mappings, similarly to \citep{mikolov13b}. For a target language $f$, we first extract a small training dictionary from a phrase table that includes word-to-word alignments between English and the target language $f$. The dictionary contains English and target-language word pairs with weights: $(x_i, y_i, w_i)_{i=1,...,n}$, where $x_i$ is an English word, $y_i$ is a target-language word, and the weight $w_i = \hat{P}(x_i|y_i)$ is the relative frequency of $x_i$ given $y_i$ as extracted from the phrase table. 

Suppose we have monolingual word embeddings for English and the target language $f$. Let $\mathbf{u_i} \in \mathcal{R}^{d_1}$ be the vector representation for English word $x_i$, $\mathbf{v_i} \in \mathcal{R}^{d_2}$  be the vector representation for target-language word $y_i$. We find a linear mapping $\mathbf{M}_{f\rightarrow e}$ by solving the following weighted least squares problem where the dictionary is used as the training data:
\begin{equation} \label{eq:cross-lingual-projection}
\mathbf{M}_{f\rightarrow e} = \arg\min_{\mathbf{M}} \sum_{i=1}^{n}w_i|| \mathbf{u_i} - \mathbf{M}\mathbf{v_i}||^2
\end{equation}

In (\ref{eq:cross-lingual-projection}) we generalize the formulation in \cite{mikolov13b} by adding frequency weights to the word pairs, so that more frequent pairs are of higher importance. Using $\mathbf{M}_{f\rightarrow e}$, for any new word in $f$ with vector representation $\mathbf{v}$, we can project it into the English vector space as the vector $\mathbf{M}_{f\rightarrow e}\mathbf{v}$.

The training dictionary plays a key role in finding an effective cross-lingual embedding mapping. To control the size of the dictionary, we only include word pairs with a minimum frequency threshold. We set the threshold to obtain approximately 5K to 6K unique word pairs for a target language, as our experiments show that larger-size dictionaries might harm the performance of representation projection for direct NER model transfer.

\subsection{Direct NER Model Transfer}

The source (English) NER system is a neural network model (with architecture NN1 or NN2) that uses only word embedding features (embeddings of a word and its surrounding context) in the English vector space. Model transfer is achieved simply by projecting the target language word embeddings into the English vector space and decoding these using the English NER system. 

More specifically, given the word embeddings of a sequence of words in a target language $f$, $(\mathbf{v}_1,...,\mathbf{v}_t)$, we project them into the English vector space by applying the linear mapping $\mathbf{M}_{f\rightarrow e}$: $(\mathbf{M}_{f\rightarrow e}\mathbf{v}_1,...,\mathbf{M}_{f\rightarrow e}\mathbf{v}_t)$. The English NER system is then applied on the projected input to produce NER tags. Words not in the target-language vocabulary are projected into their English embeddings if they are found in the English vocabulary, or into an NER-trained UNK vector otherwise.

\section{Co-Decoding}

Given two weakly supervised NER systems which are trained with different data using different models (MEMM model for annotation projection and neural network model for representation projection), we would like to design a \emph{co-decoding} scheme that can combine the outputs (views) of the two systems to produce an output that is more accurate than the outputs of individual systems.

Since both systems are statistical models and can produce confidence scores (probabilities), a natural co-decoding scheme is to compare the confidence scores of the NER tags generated by the two systems and select the tags with higher confidences scores. However, confidence scores of two weakly supervised systems may not be directly comparable, especially when comparing O tags with non-O tags (i.e., entity tags). We consider an \textit{exclude-O confidence-based co-decoding scheme} which we find to be more effective empirically. It is similar to the pure confidence-based scheme, with the only difference that it always prefers a non-O tag of one system to an O tag of the other system, regardless of their confidence scores.

In our experiments we find that the annotation projection system tends to have a high precision and low recall, i.e., it detects fewer entities, but for the detected entities the accuracy is high. The representation projection system tends to have a more balanced precision and recall. Based on this observation, we develop the following \textit{rank-based co-decoding scheme} that gives higher priority to the high-precision annotation projection system: 
\begin{itemize}
\item[1.] The combined output includes all the entities detected by the annotation projection system.

\item[2.]  It then adds all the entities detected by the representation projection system that do not conflict\footnote{Two entities detected by two different systems conflict with each other if either 1) the two entities have different spans but overlap with each other; or 2) the two entities have the same span but with different NER tags.} with entities detected by the annotation projection system (to improve recall).
\end{itemize}

Note that an entity X detected by the representation projection system does not conflict with the annotation projection system if the annotation projection system produces O tags for the entire span of X. For example, suppose the output tag sequence of annotation projection is (B-PER,O,O,O,O), of representation projection is (B-ORG,I-ORG,O,B-LOC,I-LOC), then the combined output under the rank-based scheme will be (B-PER,O,O,B-LOC,I-LOC).

\section{Experiments}

\begin{table}
\small\centering
\begin{center}
\begin{tabular}{|c|c|c|c|}
\hline   \textbf{Japanese}  & \textbf{P} & \textbf{R} & $\mathbf{F_1}$ \\
\hline      Annotation-Projection (AP) & 69.9 & 43.2 & 53.4 \\
\hline Representation-Projection (NN1) & 71.5 & 36.6 & 48.4 \\
\hline Representation-Projection (NN2) & 59.9 & 42.4 & 49.7 \\
\hline Co-Decoding (Conf): AP+NN1 & 65.7 & 49.5 & 56.5 \\
\hline Co-Decoding (Rank): AP+NN1 & 68.3 & 51.6 & \textbf{58.8} \\
\hline Co-Decoding (Conf): AP+NN2 & 59.5 & 53.3 & 56.2 \\
\hline Co-Decoding (Rank): AP+NN2 & 61.6 & 54.5 & 57.8 \\
\hline \textit{Supervised (272K)} & \textit{84.5} & \textit{80.9} & \textit{82.7}  \\
\hline
\end{tabular}

\begin{tabular}{|c|c|c|c|}
\hline   \textbf{Korean}  & \textbf{P} & \textbf{R} & $\mathbf{F_1}$ \\
\hline      Annotation-Projection (AP) & 69.5 & 26.8 & 38.7 \\
\hline Representation-Projection (NN1) & 66.1 & 23.2 & 34.4 \\
\hline Representation-Projection (NN2) & 68.5 & 43.4 & 53.1 \\
\hline Co-Decoding (Conf): AP+NN1 & 68.2 & 41.0 & 51.2 \\
\hline Co-Decoding (Rank): AP+NN1 & 71.3 & 42.8 & 53.5 \\
\hline Co-Decoding (Conf): AP+NN2 & 68.9 & 53.4 & 60.2 \\
\hline Co-Decoding (Rank): AP+NN2 & 70.0 & 53.3 & \textbf{60.5} \\
\hline \textit{Supervised (97K)} & \textit{88.2} & \textit{74.0} & \textit{80.4} \\
\hline
\end{tabular}

\begin{tabular}{|c|c|c|c|}
\hline   \textbf{German}  & \textbf{P} & \textbf{R} & $\mathbf{F_1}$ \\
\hline     Annotation-Projection (AP) & 76.5 & 60.5 & 67.5 \\
\hline Representation-Projection (NN1) & 69.0 & 48.8 & 57.2 \\
\hline Representation-Projection (NN2) & 63.7 & 66.1 & 64.9 \\
\hline Co-Decoding (Conf): AP+NN1 & 68.5 & 61.7 & 64.9 \\
\hline Co-Decoding (Rank): AP+NN1 & 72.7 & 65.0 & 68.6 \\
\hline Co-Decoding (Conf): AP+NN2 & 64.7 & 71.3 & 67.9 \\
\hline Co-Decoding (Rank): AP+NN2 & 67.1 & 72.6 & \textbf{69.7} \\
\hline \textit{Supervised (125K)} & \textit{77.8} & \textit{68.1} & \textit{72.6}  \\
\hline
\end{tabular}

\begin{tabular}{|c|c|c|c|}
\hline   \textbf{Portuguese}  & \textbf{P} & \textbf{R} & $\mathbf{F_1}$ \\
\hline      Annotation-Projection (AP) & 84.0 & 50.1 & 62.7 \\
\hline Representation-Projection (NN1) & 70.5 & 47.6 & 56.8 \\
\hline Representation-Projection (NN2) & 66.0 & 63.4 & 64.7\\
\hline Co-Decoding (Conf): AP+NN1 & 72.0 & 55.8 & 62.9 \\
\hline Co-Decoding (Rank): AP+NN1 & 77.5 & 59.7 & 67.4 \\
\hline Co-Decoding (Conf): AP+NN2 & 68.1 & 67.1 & 67.6 \\
\hline Co-Decoding (Rank): AP+NN2 & 70.9 & 68.3 & \textbf{69.6} \\
\hline \textit{Supervised (173K)} & \textit{79.8} & \textit{71.9} & \textit{75.6}  \\
\hline
\end{tabular}

\end{center}
\caption{In-house NER data: Precision, Recall and $F_1$ score on exact phrasal matches. The highest $F_1$ score among all the weakly supervised approaches is shown in bold. Same for Tables \ref{table:evaluation-conll-dev} and \ref{table:evaluation-conll-test}.} \label{table:evaluation-in-house-named}
\end{table}

In this section, we evaluate the performance of the proposed approaches for cross-lingual NER, including the 2 projection-based approaches and the 2 co-decoding schemes for combining them:
\\
(1) The annotation projection (AP) approach with heuristic data selection;
\\
(2) The representation projection approach (with two neural network architectures NN1 and NN2);
\\
(3) The exclude-O confidence-based co-decoding scheme; 
\\
(4) The rank-based co-decoding scheme.

\subsection{NER Data Sets}

We have used various NER data sets for evaluation. The first group includes in-house human-annotated newswire NER data for four languages: Japanese, Korean, German and Portuguese, annotated with over 50 entity types. The main motivation of deploying such a fine-grained entity type set is to build cognitive question answering applications on top of the NER systems. The entity type set has been engineered to cover many of the frequent entity types that are targeted by naturally-phrased questions. The sizes of the test data sets are ranging from 30K to 45K tokens.

The second group includes open human-annotated newswire NER data for Spanish, Dutch and German from the CoNLL NER data sets \cite{sang02,sang03}. The CoNLL data have 4 entity types: PER (persons), ORG (organizations), LOC (locations) and MISC (miscellaneous entities). The sizes of the development/test data sets are ranging from 35K to 70K tokens. The development data are used for tuning the parameters of learning methods.

\subsection{Evaluation for In-House NER Data}

In Table \ref{table:evaluation-in-house-named}, we show the results of different approaches for the in-house NER data. For annotation projection, the source (English) NER system is a linear-chain CRF model trained with 328K tokens of human-annotated English newswire data. The target-language NER systems are 2nd-order MEMM models trained with 1.3M, 1.5M, 2.6M and 1.5M tokens of projection-labeled data for Japanese, Korean, German and Portuguese, respectively. The projection-labeled data are selected using the heuristic data selection scheme (see Table \ref{table:data-selection-improvement}). For representation projection, the source (English) NER systems are neural network models with architectures NN1 and NN2 (see Figure \ref{figure:nn-arch}), both trained with 328K tokens of human-annotated English newswire data.

The results show that the annotation projection (AP) approach has a relatively high precision and low recall. For representation projection, neural network model NN2 (with a smoothing layer) is better than NN1, and NN2 tends to have a more balanced precision and recall. The rank-based co-decoding scheme is more effective for combining the two projection-based approaches. In particular, the rank-based scheme that combines AP and NN2 achieves the highest $F_1$ score among all the weakly supervised approaches for Korean, German and Portuguese (second highest $F_1$ score for Japanese), and it improves over the best of the two projection-based systems by 2.2 to 7.4 $F_1$ score.

We also provide the performance of \textit{supervised learning} where the NER system is trained with human-annotated data in the target language (with size shown in the bracket). While the performance of the weakly supervised systems is not as good as supervised learning, it is important to build weakly supervised systems with decent performance when supervised annotation is unavailable.  Even if supervised annotation is feasible, the weakly supervised systems can be used to pre-annotate the data, and we observed that pre-annotation can improve the annotation speed by 40\%-60\%, which greatly reduces the annotation cost.

\subsection{Evaluation for CoNLL NER Data}

\begin{table}
\small\centering
\begin{center}

\begin{tabular}{|c|c|c|c|}
\hline   \textbf{Spanish}  & \textbf{P} & \textbf{R} & $\mathbf{F_1}$ \\
\hline      Annotation-Projection (AP) & 65.5 & 59.1 & 62.1 \\
\hline Representation-Projection (NN1) & 63.9 & 52.2 & 57.4 \\
\hline Representation-Projection (NN2) & 55.3 & 51.8 & 53.5 \\
\hline Co-Decoding (Conf): AP+NN1 & 64.3 & 66.8 & \textbf{65.5}\\
\hline Co-Decoding (Rank): AP+NN1 & 63.7 & 65.3 & 64.5 \\
\hline Co-Decoding (Conf): AP+NN2 & 58.0 & 63.9 & 60.8\\
\hline Co-Decoding (Rank): AP+NN2 & 60.8 & 64.5 & 62.6 \\
\hline \textit{Supervised (264K)} & \textit{81.3} &  \textit{79.8} & \textit{80.6} \\
\hline
\end{tabular}

\begin{tabular}{|c|c|c|c|}
\hline   \textbf{Dutch}  & \textbf{P} & \textbf{R} & $\mathbf{F_1}$ \\
\hline      Annotation-Projection (AP) & 73.3 & 63.0 & 67.8 \\
\hline Representation-Projection (NN1) & 82.6 & 47.4 & 60.3 \\
\hline Representation-Projection (NN2) & 66.3 & 43.5 & 52.5 \\
\hline Co-Decoding (Conf): AP+NN1 & 72.3 & 66.5 & \textbf{69.3}\\
\hline Co-Decoding (Rank): AP+NN1 & 72.8 & 65.3 & 68.8 \\
\hline Co-Decoding (Conf): AP+NN2 & 65.3 & 64.7 & 65.0\\
\hline Co-Decoding (Rank): AP+NN2 & 69.7 & 66.0 & 67.8 \\
\hline \textit{Supervised (199K)} & \textit{82.9} &  \textit{81.7} & \textit{82.3} \\
\hline
\end{tabular}

\begin{tabular}{|c|c|c|c|}
\hline   \textbf{German}  & \textbf{P} & \textbf{R} & $\mathbf{F_1}$ \\
\hline     Annotation-Projection (AP) & 71.8 & 54.7 & 62.1 \\
\hline Representation-Projection (NN1) & 79.4 & 41.4 & 54.4 \\
\hline Representation-Projection (NN2) & 64.6 & 42.7 & 51.4 \\
\hline Co-Decoding (Conf): AP+NN1 & 70.1 & 59.5 & 64.4 \\
\hline Co-Decoding (Rank): AP+NN1 & 71.0 & 59.4 & \textbf{64.7} \\
\hline Co-Decoding (Conf): AP+NN2 & 64.2 & 59.9 & 62.0\\
\hline Co-Decoding (Rank): AP+NN2 & 66.8 & 60.6 & 63.6 \\
\hline \textit{Supervised (206K)} & \textit{81.2} & \textit{64.3} & \textit{71.8}\\
\hline
\end{tabular}

\end{center}
\caption{CoNLL NER development data.}
\label{table:evaluation-conll-dev}
\end{table}

For the CoNLL data, the source (English) NER system for annotation projection is a linear-chain CRF model trained with the CoNLL English training data (203K tokens), and the target-language NER systems are 2nd-order MEMM models trained with 1.3M, 7.0M and 1.2M tokens of projection-labeled data for Spanish, Dutch and German, respectively. The projection-labeled data are selected using the heuristic data selection scheme, where the threshold parameters $q$ and $n$ are determined via coordinate search based on the CoNLL development sets. Compared with no data selection, the data selection scheme improves the annotation projection approach by 2.7/2.0/2.7 $F_1$ score on the Spanish/Dutch/German development data. In addition to standard NER features such as $n$-gram word features, word type features, prefix and suffix features, the target-language NER systems also use the multilingual Wikipedia entity type mappings developed in \cite{ni16} to generate dictionary features and as decoding constraints, which improve the annotation projection approach by 3.0/5.4/7.9 $F_1$ score on the Spanish/Dutch/German development data.

For representation projection, the source (English) NER systems are neural network models (NN1 and NN2) trained with the CoNLL English training data. Compared with the standard CBOW word2vec model, the concatenated variant improves the representation projection approach (NN1) by 8.9/11.4/6.8 $F_1$ score on the Spanish/Dutch/German development data, as well as by 2.0 $F_1$ score on English. In addition, the frequency-weighted cross-lingual word embedding projection (\ref{eq:cross-lingual-projection}) improves the representation projection approach (NN1) by 2.2/6.3/3.7 $F_1$ score on the Spanish/Dutch/German development data, compared with using uniform weights on the same data. We do observe, however, that using uniform weights when keeping only the most frequent translation of a word instead of all word pairs above a threshold in the training dictionary, leads to performance similar to that of the frequency-weighted projection.

In Table \ref{table:evaluation-conll-dev} we show the results for the CoNLL development data. For representation projection, NN1 is better than NN2. Both the annotation projection approach and NN1 tend to have a high precision. In this case, the exclude-O confidence-based co-decoding scheme that combines AP and NN1 achieves the highest $F_1$ score for Spanish and Dutch (second highest $F_1$ score for German), and improves over the best of the two projection-based systems by 1.5 to 3.4 $F_1$ score.

In Table \ref{table:evaluation-conll-test} we compare our top systems  (confidence or rank-based co-decoding of AP and NN1, determined by the development data) with the best results of the cross-lingual NER approaches proposed in \newcite{tackstrom12}, \newcite{nothman13} and \newcite{tsai16} on the CoNLL test data. Our systems outperform the previous state-of-the-art approaches, closing more of the gap to supervised learning.

\begin{table}
\small\centering
\begin{center}

\begin{tabular}{|c|c|c|c|}
\hline   \textbf{Spanish}  & \textbf{P} & \textbf{R} & $\mathbf{F_1}$ \\
\hline \newcite{tackstrom12} & x & x & 59.3 \\
\hline \newcite{nothman13} & x & x & 61.0 \\
\hline \newcite{tsai16} & x & x & 60.6 \\
\hline Co-Decoding (Conf): AP+NN1 & 64.9 & 65.2 & \textbf{65.1}\\
\hline Co-Decoding (Rank): AP+NN1 & 64.6 & 63.9 & 64.3 \\
\hline \textit{Supervised (264K)} & \textit{82.5} & \textit{82.3} & \textit{82.4} \\
\hline
\end{tabular}

\begin{tabular}{|c|c|c|c|}
\hline   \textbf{Dutch}  & \textbf{P} & \textbf{R} & $\mathbf{F_1}$ \\
\hline \newcite{tackstrom12} & x & x & 58.4 \\
\hline \newcite{nothman13} & x & x & 64.0 \\
\hline \newcite{tsai16} & x & x & 61.6 \\
\hline Co-Decoding (Conf): AP+NN1 & 69.1 & 62.0 & \textbf{65.4}\\
\hline Co-Decoding (Rank): AP+NN1 & 69.3 & 61.0 & 64.8 \\
\hline \textit{Supervised (199K)} & \textit{85.1} &  \textit{83.9} & \textit{84.5} \\
\hline
\end{tabular}

\begin{tabular}{|c|c|c|c|}
\hline   \textbf{German}  & \textbf{P} & \textbf{R} & $\mathbf{F_1}$ \\
\hline \newcite{tackstrom12} & x & x & 40.4 \\
\hline \newcite{nothman13} & x & x & 55.8 \\
\hline \newcite{tsai16} & x & x & 48.1 \\
\hline Co-Decoding (Conf): AP+NN1 & 68.5 & 51.0 & \textbf{58.5}\\
\hline Co-Decoding (Rank): AP+NN1 & 68.3 & 50.4 & 58.0 \\
\hline \textit{Supervised (206K)} & \textit{79.6} & \textit{65.3} & \textit{71.8}\\
\hline
\end{tabular}

\end{center}
\caption{CoNLL NER test data.}
\label{table:evaluation-conll-test}
\end{table}

\section{Related Work}

The traditional annotation projection approaches \cite{yarowsky01,zitouni08,ehrmann11} project NER tags across language pairs using parallel corpora or translations. \newcite{wang14} proposed a variant of annotation projection which projects expectations of tags and uses them as constraints to train a model based on generalized expectation criteria. Annotation projection has also been applied to several other cross-lingual NLP tasks, including word sense disambiguation \cite{diab02}, part-of-speech (POS) tagging \cite{yarowsky01} and dependency parsing \cite{rasooli15}.

Wikipedia has been exploited to generate weakly labeled multilingual NER training data. The basic idea is to first categorize Wikipedia pages into entity types, either based on manually constructed rules that utilize the category information of Wikipedia \cite{richman08} or Freebase attributes \cite{alrfou15}, or via a classifier trained with manually labeled Wikipedia pages \cite{nothman13}. Heuristic rules are then developed in these works to automatically label the Wikipedia text with NER tags. \newcite{ni16} built high-accuracy, high-coverage multilingual Wikipedia entity type mappings using weakly labeled data and applied those mappings as decoding constrains or dictionary features to improve multilingual NER systems.

For direct NER model transfer, \newcite{tackstrom12} built cross-lingual word clusters using monolingual data in source/target languages and aligned parallel data between source and target languages. The cross-lingual word clusters were then used to generate universal features. \newcite{tsai16} applied the cross-lingual wikifier developed in \cite{tsai16b} and multilingual Wikipedia dump to generate language-independent labels (FreeBase types and Wikipedia categories) for $n$-grams in a document, and those labels were used as universal features.

Different ways of obtaining cross-lingual embeddings have been proposed in the literature. One approach builds monolingual representations separately and then brings them to the same space typically using a seed dictionary \cite{mikolov13b,faruqui14}. Another line of work builds inter-lingual representations simultaneously, often by generating mixed language corpora using the supervision at hand (aligned sentences, documents, etc.) \cite{vulic15,gouws15a}. We opt for the first solution in this paper because of its flexibility: we can map all languages to English rather than requiring separate embeddings for each language pair. Additionally we are able to easily add a new language without any constraints on the type of data needed. Note that although we do not specifically create inter-lingual representations, by training mappings to the common language, English, we are able to map words in different languages to a common space. Similar approaches for cross-lingual model transfer have been applied to other NLP tasks such as document classification \cite{klementiev12}, dependency parsing \cite{guo15} and POS tagging \cite{gouws15b}.

\section{Conclusion}

In this paper, we developed two weakly supervised approaches for cross-lingual NER based on effective annotation and representation projection. We also designed two co-decoding schemes that combine the two projection-based systems in an intelligent way. Experimental results show that the combined systems outperform three state-of-the-art cross-lingual NER approaches, providing a strong baseline for building cross-lingual NER systems with no human annotation in target languages.


\bibliography{ner}
\bibliographystyle{acl_natbib}

\end{document}